\documentclass[runningheads]{llncs}
\usepackage[T1]{fontenc}
\usepackage{graphicx}
\usepackage{float}
\usepackage[noadjust]{cite}
\usepackage{lipsum}
\usepackage{subcaption}
\usepackage{amsmath}
\usepackage{amssymb}
\usepackage{booktabs,makecell, multirow, tabularx} 

\begin{document}
\title{HydroVision: LiDAR-Guided Hydrometric Prediction with Vision Transformers and Hybrid Graph Learning}

%
\titlerunning{HydroVision: LiDAR-Guided Hydrometric
Prediction }
%

\author{Naghmeh Shafiee Roudbari, Ursula Eicker, \\Charalambos Poullis, Zachary Patterson\\
    \textit{Gina Cody School of Engineering and Computer Science}
    \institute{Concordia University, Montreal, QC, Canada}
    }

\authorrunning{N. Shafiee Roudbari et al.}

%
\maketitle              
\begin{abstract}
Hydrometric forecasting is crucial for managing water resources, flood prediction, and environmental protection. Water stations are interconnected, and this connectivity influences the measurements at other stations. However, the dynamic and implicit nature of water flow paths makes it challenging to extract a priori knowledge of the connectivity structure. We hypothesize that terrain elevation significantly affects flow and connectivity. To incorporate this, we use LiDAR terrain elevation data encoded through a Vision Transformer (ViT). The ViT, which has demonstrated excellent performance in image classification by directly applying transformers to sequences of image patches, efficiently captures spatial features of terrain elevation. To account for both spatial and temporal features, we employ GRU blocks enhanced with graph convolution, a method widely used in the literature. We propose a hybrid graph learning structure that combines static and dynamic graph learning. A static graph, derived from transformer-encoded LiDAR data, captures terrain elevation relationships, while a dynamic graph adapts to temporal changes, improving the overall graph representation. We apply graph convolution in two layers through these static and dynamic graphs. Our method makes daily predictions up to 12 days ahead. Empirical results from multiple water stations in Quebec demonstrate that our method significantly reduces prediction error by an average of 10\% across all days, with greater improvements for longer forecasting horizons. 


\end{abstract}
\section{Introduction}
Hydrometric forecasting is a critical component of water resource management, with significant implications for public safety, economic stability, and environmental conservation. Among the various aspects of hydrometric forecasting, water level prediction stands out due to its direct and widespread impact. Accurate forecasts provide early warnings for flood prevention, ensure the integrity of dams and bridges, and optimize water distribution for agricultural, industrial, and domestic use. Additionally, water level forecasting is essential for maintaining aquatic ecosystems and biodiversity.





Water systems exhibit temporal variability influenced by seasonal patterns, weather events, and long-term climate trends. These temporal dynamics shape the flow rates, water levels, and overall hydrological behavior over different time scales. Simultaneously, understanding spatial interactions is crucial for predicting how changes in one location propagate throughout the hydrological system. Therefore, forecasting water flow in hydrological systems inherently poses a spatiotemporal forecasting challenge, as it necessitates capturing both the temporal variability and spatial interconnections within the hydrological network to provide accurate predictions and insights into water system dynamics.

The spatial correlation of water systems is influenced by terrain elevation changes, which determine how water flows through a landscape. This parameter is crucial as it affects the speed and direction of runoff, with steeper slopes leading to faster runoff and potentially higher water levels in lower areas. It also affects the accumulation and distribution of water across different regions, contributing to the overall dynamics of the water system. 

Most papers in this area focus on historical water level data and primarily consider only the temporal correlations. Various statistical and machine learning models, such as autoregressive integrated moving average (ARIMA) \cite{bazrafshan2015hydrological}, support vector machine (SVM) \cite{asefa2006multi}, and artificial neural network (ANN) \cite{aichouri2015river}, have been widely used for this purpose. However, these methods often fall short in capturing the spatial interactions and complex dependencies within hydrological systems. Some recent work \cite{roudbari2023transglow,bai2023graph} have adopted graph neural network (GNN) based approaches to address these limitations. GNN methods excel in capturing the spatial relationships among multiple water stations, providing a more comprehensive understanding of hydrological dynamics. Despite their advantages, GNN methods still face challenges in accurately modeling the effects of terrain elevation on water flow patterns. In this study, we address the influence of terrain elevation on water level and connectivity in hydrometric forecasting. To incorporate this essential factor, we utilize LiDAR terrain elevation data encoded through a Vision Transformer (ViT). The use of Vision Transformer (ViT) stems from transformers' success in natural language processing tasks \cite{kalyan2021ammus} and, more recently, in computer vision by directly applying transformers to image patches \cite{dosovitskiy2020image}. This approach allows us to understand how variations in terrain elevation affect water flow patterns across different regions, providing a robust foundation for our forecasting model.

To model both temporal dependencies and spatial relationships among water stations, we employ Gated Recurrent Unit (GRU) blocks enhanced with graph convolution as successfully used in the recent literature\cite{zhang2018gaan,cui2020learning,roudbari2023transglow}. GRU blocks are well-suited for capturing sequential dependencies in time-series data, such as water flow measurements \cite{gharehbaghi2022groundwater}. By integrating graph convolution, which models spatial dependencies through graph structures where nodes represent water stations and edges denote relationships, we extend our model's capability to capture complex interactions in hydrological systems. Furthermore, we propose a novel hybrid graph learning structure that combines static and dynamic graph learning. Static graphs, derived from transformer-encoded LiDAR data, capture terrain elevation relationships that remain consistent over time. In contrast, dynamic graphs adapt to temporal changes in water flow patterns and connectivity between stations, thereby improving the overall graph representation and adaptability of the model. In this paper, we present the following contributions:
\begin{itemize}
  \item Incorporating Terrain Elevation for Water Level Forecasting: We integrate LiDAR-derived terrain elevation data into our hydrometric forecasting model, acknowledging its critical impact on water flow and connectivity.

 \item Proposing a Hybrid Graph Learning Structure: We introduce a novel hybrid graph learning structure that combines static and dynamic graph learning. Static graphs, derived from transformer-encoded LiDAR data, capture terrain elevation relationships, while dynamic graphs adapt to temporal changes, enhancing the overall graph representation.

 \item Demonstrating Superior Performance in Experiments: Through experiments conducted on water stations in Quebec, using the data from Environment and Natural Resources of Canada \cite{levelData2024}, our method outperforms state-of-the-art methods across all prediction horizons and performance metrics.
\end{itemize}
\section{Related Work}

Building a model that considers all the influencing parameters on the water cycle is complex due to the intricate nature of hydrological systems. Hydrometric parameters forecasting has evolved significantly over the years. Initially, statistical models such as ARIMA were the primary tools used for time series forecasting \cite{irvine1992multiplicative,papamichail2001seasonal,montanari1997fractionally}. ARIMA models are relatively easy to implement and interpret, making them suitable for short-term forecasting. However, ARIMA models have notable limitations, particularly in handling non-linear relationships and complex temporal patterns, which are often present in hydrometric data. Machine learning models like Support Vector Machines (SVM) \cite{sapankevych2009time}, Artificial Neural Networks (ANN) \cite{jain2007hybrid}, and Radial Basis Function (RBF) networks \cite{moradkhani2004improved} emerged as alternatives to overcome the limitations of statistical models. SVMs have been used for their robustness in handling non-linear relationships. ANNs, particularly feedforward neural networks, have been widely applied due to their ability to approximate any continuous function, offering greater flexibility than ARIMA models. However, ANN models often require large amounts of training data and are prone to overfitting. RBF networks, a variant of ANNs, provide better generalization capabilities but still face challenges in capturing long-term dependencies in time series data. The advent of Recurrent Neural Networks (RNNs) \cite{zhang1998time} marked a significant advancement in time series forecasting, particularly for sequential data. RNNs and their variants like Long Short-Term Memory (LSTM) networks \cite{hochreiter1997long} are specifically designed to capture long-term dependencies and temporal correlations in data, addressing the limitations of traditional ANN models. However, RNNs and LSTMs struggle with spatial dependencies, prompting the integration of Convolutional Neural Networks (CNNs) and Graph Neural Networks (GNNs). CNNs, known for their effectiveness in capturing spatial relationships, have been applied to hydrometric forecasting by modeling spatial dependencies among multiple stations \cite{atashi2023comparative}. GNNs further enhance this capability by operating on graph-structured data, capturing complex spatial relationships and interactions within the hydrological system \cite{chen2021flood}. 

To address the limitations of both spatial and temporal modeling, recent work \cite{shang2021discrete,roudbari2022simpler,wu2020connecting} have proposed hybrid models such as Graph Convolutional Recurrent Networks (GCRNs). GCRNs combine the strengths of GRU blocks for sequential data processing with graph convolution operations to capture spatial dependencies, providing a comprehensive approach to spatiotemporal forecasting. These hybrid models, inspired by the success of both RNNs and GNNs in their respective domains, offer a more robust framework for hydrometric forecasting. However, even these advanced models have shortcomings, particularly in incorporating complex domain information. Therefore, we propose integrating terrain elevation through Vision Transformers (ViTs) to create static graph structures to enhance prediction.

Transformers, initially introduced for machine translation tasks \cite{vaswani2017attention}, have revolutionized natural language processing (NLP) tasks. These models have been pretrained on vast amounts of text and fine-tuned for various applications. Notable examples include BERT \cite{devlin2018bert} and the GPT series \cite{floridi2020gpt}, which uses language modeling for pretraining. Drawing inspiration from the success of Transformers in NLP, Vision Transformers (ViT) have emerged as a powerful tool in computer vision. The original ViT \cite{dosovitskiy2020image} divides images into patches, applies Multi-Head Attention (MHA) \cite{vaswani2017attention} to these patches, and uses a learnable classification token to capture a global visual representation, enabling effective image classification.

\section{Dataset}

In this work, we utilize two types of data: LiDAR data and time series water level data, as outlined below. These datasets provide complementary information that enhances our water level forecasting model.

\subsection{LiDAR Data}
\begin{figure*}[!t]
    \centering
    \begin{subfigure}{0.3\textwidth}
        \includegraphics[width=\textwidth]{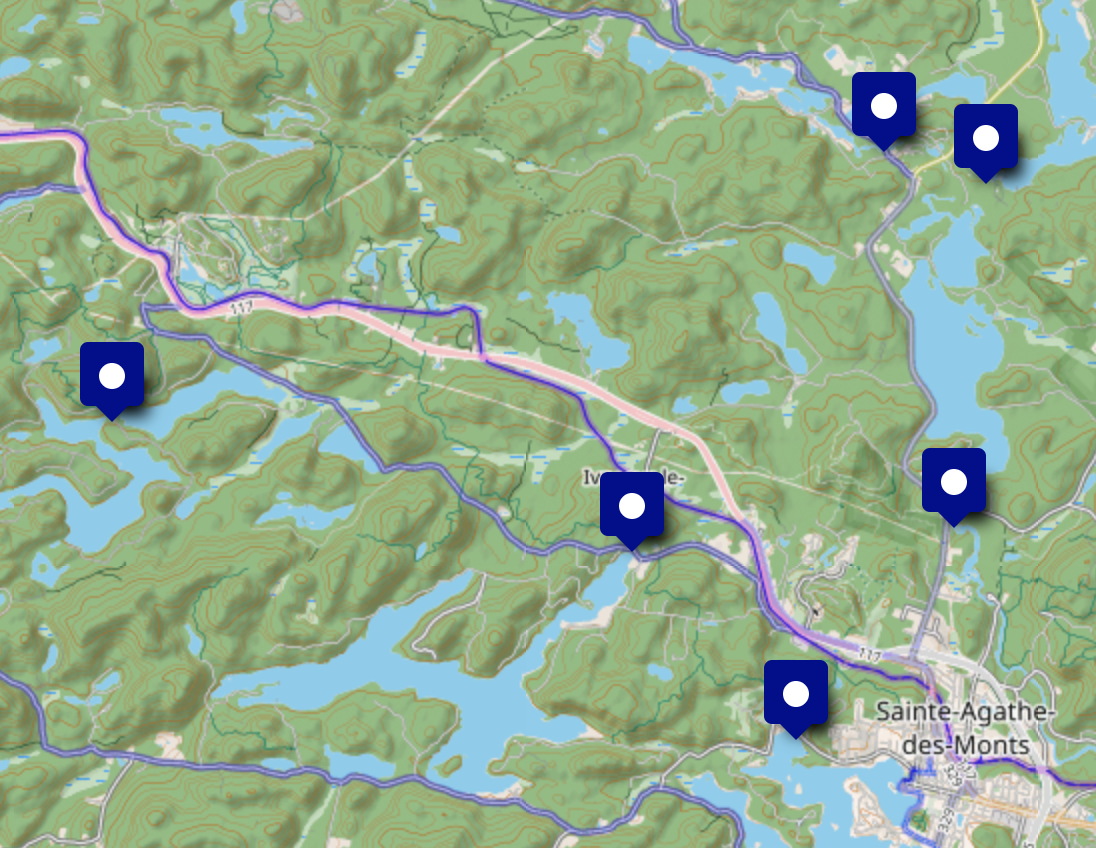}
        \caption{}
        \label{subfig:station-map}
    \end{subfigure}
    \hfill
    \begin{subfigure}{0.35\textwidth}
        \includegraphics[width=\textwidth]{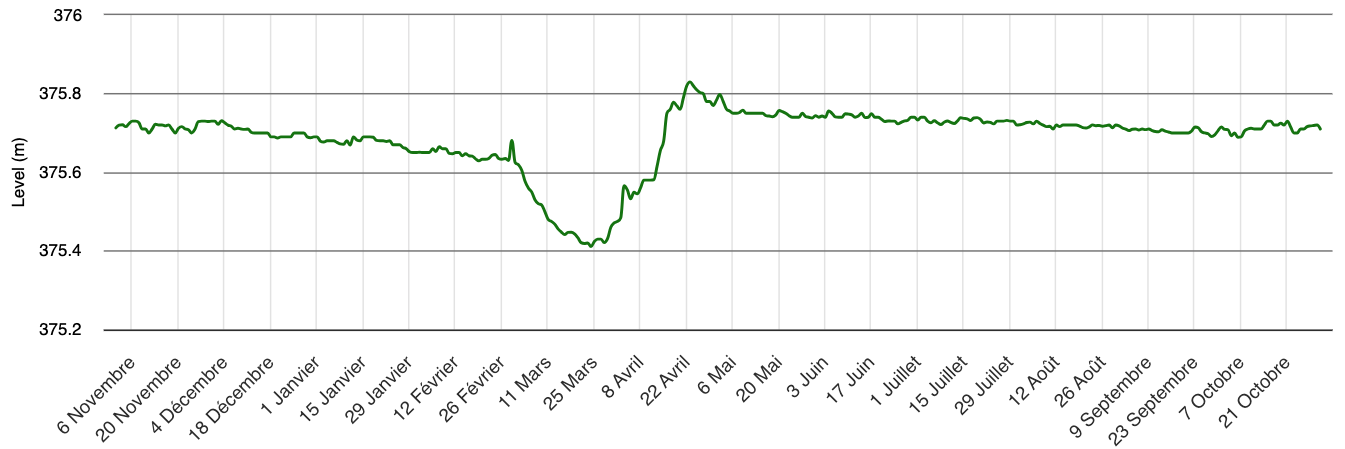}
        \caption{}
        \label{subfig:variation-plot}
    \end{subfigure}
    \hfill
    \begin{subfigure}{0.3\textwidth}
        \includegraphics[width=\textwidth]{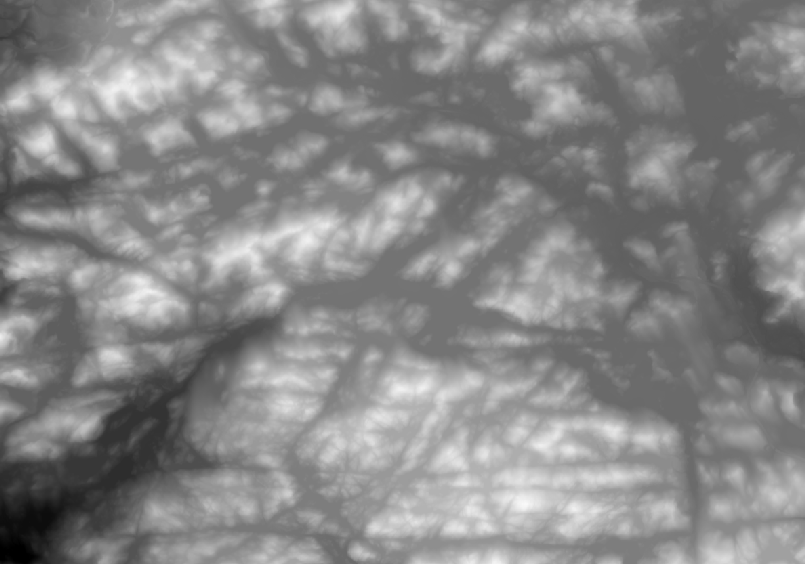}
        \caption{}
        \label{subfig:lidar-map}
    \end{subfigure}
    \caption{(a) The geographic distribution of water level monitoring stations around the Sainte-Agathe-des-Monts (b) Variation in water levels throughout the year at Lake Papineau in Sainte-Agathe-des-Monts (c) Visualization of the Digital Terrain Model (DTM) near Sainte-Agathe-des-Monts }
    \label{fig:temporal_plot}
\end{figure*}

The LiDAR data, provided by the Ministère des Ressources naturelles et des Forêts (MRNF) \cite{MRNF2024} as part of the provincial LiDAR sensor data acquisition project, includes the Digital Terrain Model (DTM). This DTM is a raster file with a spatial resolution of 1 meter, providing precise numerical values representing altitudes in meters relative to mean sea level. Elevation values are derived through linear interpolation across an irregular triangle network created from ground points. The DTM images are produced by superimposing the Digital Elevation Model (DEM) with the shaded DEM to accentuate relief, using color gradients and transparency. With its high spatial resolution, it is also extensively used in creating hydrological models, planning road construction, managing flood risks, and conducting visual landscape analyses. Figure \ref{subfig:lidar-map} visualizes the DTM of the study area.

\subsection{Timeseries data}The time series data consists of daily water level measurements from six stations on bodies of water in a specific region in Quebec, spanning 40 years from 1981 to 2021. Provided by Environment and Climate Change Canada \cite{levelData2024}, this dataset is crucial for understanding water level variations over time. Missing values in the data are replaced by a weighted average of the previous and next year's data. Figures \ref{subfig:station-map} and \ref{subfig:variation-plot} illustrate the station coverage on the map and the variation of water levels at one station over a year, respectively. 

To ensure that computational resources are efficiently utilized and data processing remains manageable, we have selected the closest geographically clustered monitoring stations from all available stations scattered across a wide region. This selection is due to the necessity of loading the LiDAR data that covers the entire study area.

\section{Objective}

Given the LiDAR data, our goal is to uncover the underlying spatial relationships. These spatial relationships will then be used as inputs for another function, combined with 2D time series data from \( n \) stations over \( m \) timestamps, to predict future values.

Let \( L \) represent the LiDAR data, and \( T \) represent the time series data, where \(T \in \mathbb{R}^{n \times m} \).We aim to find a function \( f \) that captures the spatial relationships from \( L \):
\[
f: L \rightarrow S
\]
where \( S \) represents the spatial relationships. Next, we define another function \( g \) that takes \( S \) and \( T \) as inputs to predict future values \( \hat{T} \):
\[
g: (S, T) \rightarrow \hat{T}
\]

The overall objective is to find the optimal functions \( f \) and \( g \) such that the predicted future values \( \hat{T} \) closely match the actual future values.

\section{Methodology}
\begin{figure}
    \centering
    \includegraphics[width=\linewidth]{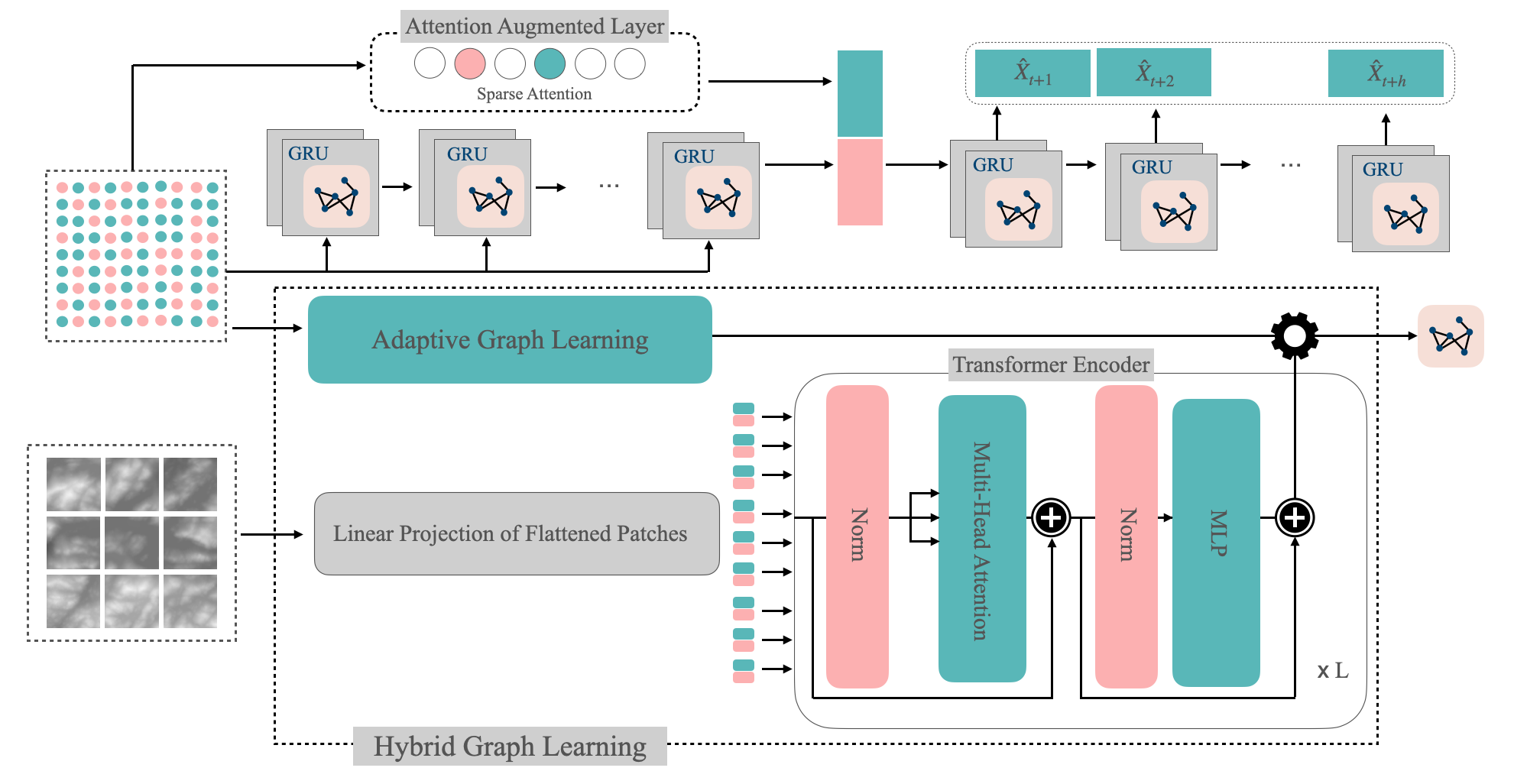}
    \caption{Main Architecture of the HydroVision Framework. The architecture integrates LiDAR data via a Vision Transformer and combines adaptive graph learning with GCRNs to capture both spatial and temporal dependencies for accurate water level forecasting.}
    \label{fig:main-arch}
\end{figure}

In this section, we introduce the HydroVision framework for water level forecasting, which integrates two essential components. First, we discuss the foundational approach, GCRN blocks. Second, we describe the hybrid graph learning layer incorporated in our study. Together, these elements constitute the HydroVision framework, as illustrated in Figure \ref{fig:main-arch}.

\subsection{Foundational Approach}

\subsubsection{Graph Convolutional Recurrent Network}
GCRN combines graph convolution operations and Gated Recurrent Units (GRUs) to tackle spatiotemporal forecasting challenges. The graph convolution captures spatial dependencies, while the GRU models temporal variability. Due to the success of previous research \cite{li2017diffusion,shang2021discrete} using this method for forecasting purposes, we adopt the GCRN formulation expressed as:

\begin{equation}
\label{eq:graph_conv_new}
{Z}^{(l)} = \sigma (\tilde{A}Z^{(l-1)}W^{(l)} + b^l)
\end{equation}

In this equation, $\tilde{A}$ is the learned adjacency matrix, $W^l$ and $b^l$ are the weight and bias matrices, and $\sigma$ is the activation function. The GRU's traditional MLP layers are replaced by this graph convolution, resulting in the following equations for the reset gate ($r^t$), update gate ($u^t$), candidate cell state ($c^t$), and hidden state ($h^t$):

\begin{equation}
\label{eq:rt_new}
r^{t} = \sigma (\mathcal{F}(\tilde{A}, [X^t, h^{t-1}]) + c_r)
\end{equation}

\begin{equation}
\label{eq:ut_new}
u^{t} = \sigma (\mathcal{F}(\tilde{A}, [X^t, h^{t-1}]) + c_u)
\end{equation}

\begin{equation}
\label{eq:ct_new}
c^{t} = \tanh (\mathcal{F}(\tilde{A}, [X^t, r^t \odot h^{t-1}]) + c_c)
\end{equation}

\begin{equation}
\label{eq:ht_new}
h^{t} = (u^t \odot h^{t-1}) + (1 - u^t) \odot c^t
\end{equation}

Here, $\mathcal{F}$ represents the graph convolution operation, $X^t$ is the input at time $t$, and $\odot$ denotes element-wise multiplication. The activation functions $\sigma$ and $\tanh$ regulate the network's internal state transitions.

\subsubsection{Encoder-Decoder with Augmented Attention }

The encoder-decoder model is an effective approach for sequence-to-sequence tasks, widely used in fields like machine translation \cite{cho2014learning} and time series forecasting \cite{shang2021discrete}. The basic encoder-decoder model can struggle with information compression, especially for longer sequences. To address this, we use the architecture with an augmented attention layer. The attention mechanism computes a weighted sum of the input sequence elements, creating an augmented hidden state $H$:

\begin{equation}
\label{eq:encoder2}
H = \text{Concat}[h^t, C]
\end{equation}

Here, $h^t$ is the final hidden state of the encoder, and $C$ is the context vector from the attention layer. This augmented state helps the decoder focus on relevant parts of the input data \cite{roudbari2023transglow}. The original attention mechanism has a quadratic computational complexity with respect to the sequence length. To mitigate this, Zhou, Haoyi, et al. \cite{zhou2021informer} proposed ProbSparse Self-attention, which selects a subset of $k$ queries based on a probability distribution:

\begin{equation}
\label{eq:prob_q}
\hat{Q} = M(Q,K),
\text{Attention}(Q,K,V) = \text{Softmax}\left(\frac{\hat{Q}K^T}{\sqrt{d}}\right)V
\end{equation}

In these equations, $Q$, $K$, and $V$ represent query, key, and value matrices, and $d$ is the dimension. The probability distribution $M$ determines the importance of each token in the sequence, including more relevant tokens in the sparse query matrix and excluding less relevant ones. This efficient attention mechanism improves the scalability of the encoder-decoder model.

\subsection{Vision Transformer}

In this study, we encode LiDAR elevation data using the Transformer model, as presented in the work by Dosovitskiy et al. \cite{dosovitskiy2020image}. The Transformer model, originally designed for natural language processing, has been adapted to handle image data, proving highly effective in tasks requiring spatial understanding and feature extraction from images.

The core idea behind using Transformers for image recognition involves dividing the input image into a sequence of patches, which are then processed by the Transformer encoder. To encode the LiDAR elevation data, we first partition the data into non-overlapping patches of size \(16 \times 16\). Each patch is then flattened into a vector and projected to a fixed-dimensional embedding.

The sequence of embedded patches, along with positional encodings \(\mathbf{E}_{pos}\), is then fed into the Transformer encoder. The positional encodings are crucial for retaining spatial information, as the Transformer architecture does not inherently capture the order of the input sequence. The positional encoding can be defined as:

\begin{equation}
\label{eq:positional_encoding}
\mathbf{E}_{pos}(pos, 2i) = \sin \left( \frac{pos}{10000^{2i/d}} \right), \quad
\mathbf{E}_{pos}(pos, 2i+1) = \cos \left( \frac{pos}{10000^{2i/d}} \right)
\end{equation}

where \(pos\) is the position, \(i\) is the dimension, and \(d\) is the embedding size.

The embedded patches and positional encodings are combined and processed through the Transformer encoder layers, which consist of multi-head self-attention and feed-forward neural networks. The output from the Transformer encoder provides a rich, context-aware representation of the LiDAR elevation data, capturing both local and global spatial features.

The process can be summarized by the following equation for the Transformer encoder layer:

\begin{equation}
\label{eq:transformer_encoder}
\mathbf{A_{elevation}} = \text{TransformerEncoder}(\mathbf{E} + \mathbf{E}_{pos})
\end{equation}

where \(\mathbf{E}\) is the sequence of embedded patches, \(\mathbf{E}_{pos}\) is the positional encoding, and \(\mathbf{G1}\) is the output of the Transformer encoder. By leveraging the Transformer model for encoding LiDAR elevation data, we can effectively capture complex spatial relationships and provide a robust input representation for subsequent processing in our HydroVision framework.

\subsection{Hybrid Graph Learning}

To adaptively learn the spatial relationships between objects, we adopt the adaptive graph generation technique as defined in \cite{bai2020adaptive}:

\begin{equation}
\label{eq:graph_conv}
A_{\text{adaptive}} = \text{softmax}(\text{ReLU}(E1 \cdot E2^T))
\end{equation}

In this equation, \(E1\) and \(E2\) represent node embeddings that are randomly initialized and subsequently learned during the training process. This method allows the model to dynamically adjust the spatial relationships between nodes based on the data. To enhance our model's performance, we integrate both the adaptively learned graph and the elevation encoded output into a combined graph representation. This can be expressed as:

\begin{equation}
\label{eq:combined_graph}
\hat{A}= \alpha A_{\text{adaptive}} + (1 - \alpha) A_{\text{elevation}}
\end{equation}

where \(\alpha\) is a weighting parameter that balances the contribution of each graph. The combined graph \(\hat{A}\) encapsulates the comprehensive underlying information used in our graph convolution operations.

\section{Experiments}
\subsection{Settings}
In our experiments, we allocate 70\% of the dataset for training, 10\% for validation, and the remaining 20\% for testing. We use a batch size of 64. Both the length of the historical sequences and the prediction horizon are set to 12 time steps. The maximum number of training epochs is capped at 300, though early stopping is employed if the validation performance does not improve for 20 consecutive epochs.

Training is performed using the Adam optimizer with the Mean Absolute Error (MAE) as the loss function. To enhance generalization, curriculum learning is applied. The initial learning rate is set to 0.01, with a decay ratio of 0.1. The attention mechanism in the network utilizes 8 heads. The model is implemented using PyTorch version 1.7.1, and all experiments are conducted on an NVIDIA GeForce RTX 2080 Ti GPU with 11GB of memory.

\subsection{Comparative Performance Evaluation}
table \ref{performance_table} provides a comparative analysis of different models for water level forecasting. The performance metrics used are Mean Absolute Error (MAE) and Root Mean Square Error (RMSE).

\begin{table}
\caption{ Comparative Performance of Various Models for Water Level Forecasting. The table displays the Mean Absolute Error (MAE) and Root Mean Square Error (RMSE) for different forecasting models across various prediction horizons (3, 6, 9, and 12 days).}
\label{performance_table}
\centering
\begin{tabular}{cc|cccc}

\hline
\multicolumn{1}{l}{Method}&
\multicolumn{1}{c}{Metric}&
\multicolumn{4}{c}{Horizon}                             \\ \hline &   &  \multicolumn{1}{c}{3 Days} &
\multicolumn{1}{c}{6 Days} & \multicolumn{1}{c}{9 Days}    & \multicolumn{1}{c}{12 Days}                                                    \\ \hline

\multirow{3}{*}{AGCRN} & \multicolumn{1}{c|}{MAE}  & \multicolumn{1}{c}{0.514} & \multicolumn{1}{c}{0.526} & \multicolumn{1}{c}{0.547}  & \multicolumn{1}{c}{0.573}\\ &\multicolumn{1}{c|}{RMSE}  & \multicolumn{1}{c}{0.841} & \multicolumn{1}{c}{0.857} & \multicolumn{1}{c}{0.886}& \multicolumn{1}{c}{0.871}  \\ \hline

\multirow{3}{*}{Informer} & \multicolumn{1}{c|}{MAE} & \multicolumn{1}{c}{0.717} & \multicolumn{1}{c}{0.733} & \multicolumn{1}{c}{0.772} & \multicolumn{1}{c}{0.825}\\ & \multicolumn{1}{c|}{RMSE} & \multicolumn{1}{c}{0.125} & \multicolumn{1}{c}{0.141} & \multicolumn{1}{c}{0.186} & \multicolumn{1}{c}{0.252}\\ \hline

\multirow{3}{*}{DCGCN}  
& \multicolumn{1}{c|}{MAE}  & \multicolumn{1}{c}{0.145}    
& \multicolumn{1}{c}{0.317}   & \multicolumn{1}{c}{0.553} & \multicolumn{1}{c}{0.796}   \\ 
& \multicolumn{1}{c|}{RMSE}  & \multicolumn{1}{c}{0.162}    
& \multicolumn{1}{c}{0.346}   & \multicolumn{1}{c}{0.611}  & \multicolumn{1}{c}{0.869}  \\ \hline

\multirow{3}{*}{STtransformer}  
& \multicolumn{1}{c|}{MAE}  & \multicolumn{1}{c}{0.055}    
& \multicolumn{1}{c}{0.068}   & \multicolumn{1}{c}{0.072} & \multicolumn{1}{c}{0.085}   \\ 
& \multicolumn{1}{c|}{RMSE}  & \multicolumn{1}{c}{0.074}    
& \multicolumn{1}{c}{0.094}   & \multicolumn{1}{c}{0.109}  & \multicolumn{1}{c}{0.128}  \\ \hline

\multirow{3}{*}{GTS}  
& \multicolumn{1}{c|}{MAE}  & \multicolumn{1}{c}{0.053}    
& \multicolumn{1}{c}{0.064}   & \multicolumn{1}{c}{0.078} & \multicolumn{1}{c}{0.099}   \\ 
& \multicolumn{1}{c|}{RMSE}  & \multicolumn{1}{c}{0.080}    
& \multicolumn{1}{c}{0.096}   & \multicolumn{1}{c}{0.113}  & \multicolumn{1}{c}{0.130}  \\ \hline

\multirow{3}{*}{STAWnet}  
& \multicolumn{1}{c|}{MAE}  & \multicolumn{1}{c}{0.043}    
& \multicolumn{1}{c}{0.048}   & \multicolumn{1}{c}{0.059} & \multicolumn{1}{c}{0.062}   \\ 
& \multicolumn{1}{c|}{RMSE}  & \multicolumn{1}{c}{0.067}    
& \multicolumn{1}{c}{0.079}   & \multicolumn{1}{c}{0.093} & \multicolumn{1}{c}{0.101}   \\ \hline

\multirow{3}{*}{MTGNN}  
& \multicolumn{1}{c|}{MAE}  & \multicolumn{1}{c}{0.039}    
& \multicolumn{1}{c}{0.047}   & \multicolumn{1}{c}{0.059} & \multicolumn{1}{c}{0.064}   \\ 
& \multicolumn{1}{c|}{RMSE}  & \multicolumn{1}{c}{0.060}    
& \multicolumn{1}{c}{0.082}   & \multicolumn{1}{c}{0.093}   & \multicolumn{1}{c}{0.106} \\ \hline

\multirow{3}{*}{Hydrovision}  
& \multicolumn{1}{c|}{MAE}  & \multicolumn{1}{c}{0.031}    
& \multicolumn{1}{c}{0.043}   & \multicolumn{1}{c}{0.050} & \multicolumn{1}{c}{0.056}    \\ 
& \multicolumn{1}{c|}{RMSE}  & \multicolumn{1}{c}{0.057}    
& \multicolumn{1}{c}{0.075}   & \multicolumn{1}{c}{0.088} & \multicolumn{1}{c}{0.097}   \\ \hline

\end{tabular}
\end{table}

AGCRN \cite{bai2020adaptive}, designed to capture both temporal and spatial dependencies within graph-structured data through RNNs and GNNs, demonstrates a pretty stable performance over different prediction horizons. This reflects its capability to handle both types of dependencies. The Informer \cite{zhou2021informer} model is an enhanced version of the Transformer architecture, indicating some struggle with extended forecasts.

DCGCN \cite{lin2023dynamic}, employing the GCRN block structure and a graph learning approach, exhibits a significant increase in errors over time. This suggests that DCGCN faces considerable difficulty with longer-term forecasts. On the other hand, the STtransformer Network \cite{xu2020spatial}, which captures spatial and temporal dependencies using transformers, maintains lower errors compared to many other models, showcasing good performance across all prediction horizons.

The GTS method \cite{shang2021discrete}, a scalable spatiotemporal forecasting approach, demonstrates strong performance with relatively low error increments over time, indicating its efficiency compared to AGCRN and Informer. Similarly, STAWnet \cite{tian2021spatial}, known for effectively capturing spatial and temporal information using advanced attention mechanisms, performs well, maintaining low errors across all horizons.

MTGNN \cite{wu2020connecting}, integrating a graph learning module with a mix-hop propagation layer and a dilated inception layer for optimal learning, shows very low errors, indicating robust performance across all prediction horizons. Finally, Hydrovision, the proposed model in this study, consistently outperforms the other models in both MAE and RMSE. Hydrovision's ability to maintain low error rates over all tested horizons underscores its superior effectiveness for water level forecasting.

It is important to note that given the domain of the dataset, which involves small variations within a meter, higher errors are particularly concerning. For instance, the errors exhibited by DCGCN, AGCRN, and Informer may not be considered good. Hydrovision, along with MTGNN and STAWnet, stands out for maintaining errors within an acceptable range, thereby proving more suitable for the precise nature of this domain.

\subsection{Ablation Study}
To evaluate the impact of incorporating the Vision Transformer (ViT) and LiDAR elevation data in our hybrid graph learning approach, we conducted an ablation study. We compared the performance of the model with the full hybrid graph learning approach, against a variant where the ViT and LiDAR elevation data were excluded from the main architecture. The results are summarized in Table \ref{tab:ablation_study}.

\begin{table}[ht]
\centering
\caption{Ablation Study Results Comparing the Effectiveness of the Hybrid Graph Learning Approach}
\label{tab:ablation_study}
\begin{tabular}{c|cccc|cccc}
\hline
\multicolumn{1}{c}{Metric} & \multicolumn{4}{c}{Hybrid Graph Leanring} & \multicolumn{4}{c}{Adaptive Graph Leanring} \\ \hline
& 3 Days & 6 Days & 9 Days & 12 Days & 3 Days & 6 Days & 9 Days & 12 Days \\ \hline
MAE & 0.031 & 0.043 & 0.050 & 0.056 & 0.034 & 0.047 & 0.057 & 0.066 \\ 
RMSE & 0.057 & 0.075 & 0.088 & 0.097 & 0.061 & 0.079 & 0.096 & 0.108 \\ \hline
\end{tabular}
\end{table}

These results clearly demonstrate that the Hybrid Graph Learning approach enhances forecasting accuracy more effectively than the Adaptive Graph Learning approach, particularly over longer prediction periods. The improvement in both MAE and RMSE indicates that integrating the hybrid approach results in a more robust and accurate forecasting model, making it a preferable choice for applications requiring extended time horizon predictions.

This analysis underscores the value of both the ViT and LiDAR data in enhancing the model's accuracy. The detailed terrain information provided by the LiDAR data, combined with the ViT's ability to capture complex spatial patterns, contributes to more precise water level forecasting, validating the effectiveness of our proposed approach.

\section{Conclusion}
In this paper, we introduced a novel approach to water level forecasting by leveraging advanced graph learning techniques and LiDAR elevation data. Our approach integrates a Hybrid Graph Learning framework with a Vision Transformer (ViT) to enhance the accuracy of water level predictions across various time horizons. 

Our study underscores the importance of high-resolution spatial data in enhancing the predictive performance of graph-based models. The LiDAR data provides detailed elevation information that enriches the spatial context of the predictions, leading to more accurate and reliable forecasts. This finding supports the value of integrating rich, domain-specific data into forecasting models to capture nuanced spatial dependencies.

Overall, the proposed Hybrid Graph Learning approach represents a significant advancement in the field of water level forecasting. It combines state-of-the-art graph convolution techniques with cutting-edge transformer models and spatially rich LiDAR data, setting a new benchmark for accuracy and reliability in this domain. Future work could explore further refinements to the model and assess its applicability to other environmental forecasting tasks, extending the benefits of advanced graph learning methods and high-resolution spatial data to a broader range of applications.

\section{Acknowledgments}

\noindent C.P. This work is financially supported by the Mathematics of Information Technology and Complex Systems' Accelerate programme under grant agreement IT29301 and the Natural Sciences and Engineering Research Council of Canada Grants RGPIN-2021-03479 (NSERC DG). 

\noindent U.E. This research was undertaken, in part, thanks to funding from the Canada Excellence Research Chairs Program.

\noindent Z.P. This research was undertaken, in part, based on support from the Gina Cody School of Engineering of Concordia University FRS.

\bibliographystyle{splncs04}
\bibliography{IEEEexample}

\end{document}